\documentclass[runningheads]{llncs}
\usepackage{graphicx}
\usepackage{url}            
\usepackage{booktabs}       
\usepackage{amsfonts}       
\usepackage{nicefrac}       
\usepackage{microtype}      
\usepackage{tikz}
\usepackage{tabularx}
\usepackage{hhline}
\usepackage{adjustbox}
\usepackage{multirow}
\usepackage{mathtools}
\usepackage{bbm}
\usepackage{color, colortbl}
\usepackage[misc]{ifsym}

\newcommand\tf[1]{\textbf{#1}}

\def\ie{\textit{i.e.}}

\newcommand{\myparagraph}[1]{\vspace{1pt}\noindent{\bf{#1}}~~}

\usepackage{colortbl}
\definecolor{lightgray}{gray}{0.75}
\definecolor{lightergray}{gray}{0.85}
\definecolor{Blue}{RGB}{3, 31, 97}
\definecolor{Blue1}{RGB}{214, 235, 245}
\definecolor{Blue2}{RGB}{235, 245, 250}
\definecolor{Gray}{RGB}{247, 252, 255}

\definecolor{convcolor}{HTML}{412F8A}
\definecolor{resnetcolor}{HTML}{8DA0CB}
\definecolor{vitcolor}{HTML}{fc8e62}

\newcommand{\convcolor}[1]{\textcolor{convcolor}{#1}}

\newcommand{\cb}{\convcolor{$\bullet$\,}}
\newcommand{\gr}{\rowcolor[gray]{.95}}

\begin{document}
\title{Bootstrapping Semi-supervised Medical \\ Image Segmentation with Anatomical-aware Contrastive Distillation}
\titlerunning{Anatomical-aware Contrastive Distillation for Medical Image Segmentation}
%
\author{Chenyu You \textsuperscript{1(\Letter)} \and Weicheng Dai \inst{2} \and Yifei Min \inst{5} \and Lawrence Staib \inst{1,3,4} \and \\ James S. Duncan \inst{1,3,4,5}}
\authorrunning{C. You et al.}
\institute{\textsuperscript{1}Department of Electrical Engineering, Yale University
\\
\email{chenyu.you@yale.edu}\\
\textsuperscript{2}Department of Computer Science and Engineering, New York University
\\
\textsuperscript{3}Department of Biomedical Engineering, Yale University \\
\textsuperscript{4}Department of Radiology and Biomedical Imaging, Yale University\\
\textsuperscript{5}Department of Statistics and Data Science, Yale University\\
}

\maketitle              

\begin{abstract}
Contrastive learning has shown great promise over annotation scarcity problems in the context of medical image segmentation. Existing approaches typically assume a balanced class distribution for both labeled and unlabeled medical images. However, medical image data in reality is commonly imbalanced (\ie, multi-class label imbalance), which naturally yields blurry contours and usually incorrectly labels rare objects. Moreover, it remains unclear whether all negative samples are equally negative. In this work, we present \tf{ACTION}, an \tf{A}natomical-aware \tf{C}on\tf{T}rastive d\tf{I}stillati\tf{ON} framework, for semi-supervised medical image segmentation. Specifically, we first develop an iterative contrastive distillation algorithm by softly labeling the negatives rather than binary supervision between positive and negative pairs. We also capture more semantically similar features from the randomly chosen negative set compared to the positives to enforce the diversity of the sampled data. Second, we raise a more important question: Can we really handle imbalanced samples to yield better performance? Hence, the \textit{key innovation} in {ACTION} is to learn global semantic relationship across the entire dataset and local anatomical features among the neighbouring pixels with minimal additional memory footprint. During the training, we introduce anatomical contrast by actively sampling a sparse set of hard negative pixels, which can generate smoother segmentation boundaries and more accurate predictions. Extensive experiments across two benchmark datasets and different unlabeled settings show that ACTION significantly outperforms the current state-of-the-art semi-supervised methods.


\end{abstract}

\section{Introduction}
\label{section:intro}
Manually labeling sufficient medical data with pixel-level accuracy is time-consuming, expensive, and often requires domain-specific knowledge. To bypass the cost for labeled data, semi-supervised learning (SSL) is one of the promising, conventional ways to train models with weaker forms of supervision, given a large amount of unlabeled data. Existing SSL methods include adversarial training \cite{zhang2017deep,li2020shape,you2020unsupervised,yang2020nuset,you2022class}, deep co-training \cite{qiao2018deep,zhou2019semi}, mean teacher schemes \cite{tarvainen2017mean,yu2019uncertainty}, multi-task learning \cite{luo2020semi,kervadec2019curriculum,chen2019multi,you2022incremental}, and contrastive learning \cite{chaitanya2020contrastive,hu2021semi,you2022momentum,you2022simcvd,you2022mine,you2023rethinking}.

\begin{figure}[t]
\centering
\includegraphics[width=0.95\linewidth]{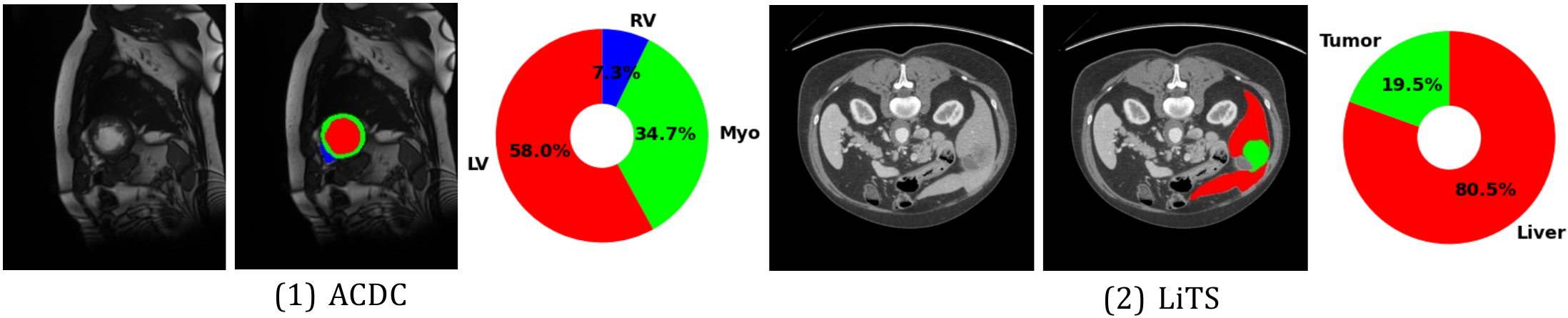}
\vspace{-10pt}
\caption{Examples of two benchmarks (\ie, ACDC and LiTS) showing
the large variations of class distribution.} 
\label{fig:distribution}
\vspace{-15pt}
\end{figure}

Among the aforementioned methods, contrastive learning \cite{he2020momentum,chen2020simple} has recently prevailed for DNNs to rich visual representations from unlabeled data. The predominant promise of label-free learning is to capture the similar semantic relationship and anatomical structure between neighboring pixels from massive unannotated data. However, going to realistic clinical scenarios will have the following shortcomings. First, different medical images share similar anatomical structures, but prior methods follow the standard contrastive learning \cite{chen2020simple,he2020momentum} in comparing positive and negative pairs by binary supervision. That naturally leads to the issues of false negatives in representation learning \cite{tejankar2021isd,huynh2022boosting}, which would hurt segmentation performance. Second, the underlying class distribution of medical image data is highly imbalanced, as illustrated in Figure \ref{fig:distribution}. It is well known that such imbalanced distribution will severely hurt the segmentation quality \cite{li2020analyzing}, which may result in blurry contours and mis-classify minority classes due to the occurrence frequencies \cite{zhu2014capturing}. That naturally questions whether contrastive learning can still work well in those imbalance scenarios.

In this work, we present a principled framework called \textbf{A}natomical-aware \textbf{C}on\textbf{T}rastive d\textbf{I}stillati\textbf{ON} (\textbf{ACTION}), for multi-class medical image segmentation. In contrast to prior work \cite{chaitanya2020contrastive,hu2021semi,you2022simcvd} which directly distinguish two image samples of the similar anatomical features that are in the negative pairs, the \textbf{key innovation} in ACTION is to actively learn more balanced representations by \textit{dynamically} selecting samples that are semantically similar to the queries, and contrasting the model's \textit{anatomical-level} features with the target model's in \tf{imbalanced} and \tf{unlabeled} clinical scenarios. Specifically, we introduce two strategies to improve overall segmentation quality: (1) we believe that all negative samples are not equally negative. Thus, we propose relaxed contrastive learning by using soft labeling on the negatives. In other words, we randomly sample a set of image samples as anchor points to ensure \tf{diversity} in the set of sampled examples. Then the teacher model predicts the underlying probability distribution over neighboring samples by computing the anatomical similarities between the query and the anchor points in the memory bank, and the student model tries to learn from the teacher model. Such a strategy is much more regularized by mincing the same neighborhood anatomical similarity to improve the quality of the anatomical features; (2) to create strong contrastive views on anatomical features, we introduce \tf{AnCo}, another new contrastive loss designed at the anatomical level, by sampling a set of pixel-level representation as queries, and pulling them closer to the mean feature of all representations in a class (positive keys), and pulling other representations apart from other class (negative keys). In addition to reducing the high memory footprint and computation complexity, we use active sampling to dynamically select a sparse set of queries and keys during the training. We apply ACTION on two benchmark datasets under different unlabeled settings. Our experiments show that ACTION can dramatically outperform the state-of-the-art SSL methods. We believe that our proposed ACTION can be a strong baseline for the related medical image analysis tasks in the future.
\section{Method}
\label{section:method}

\begin{figure}[t]
\centering
\includegraphics[width=\linewidth]{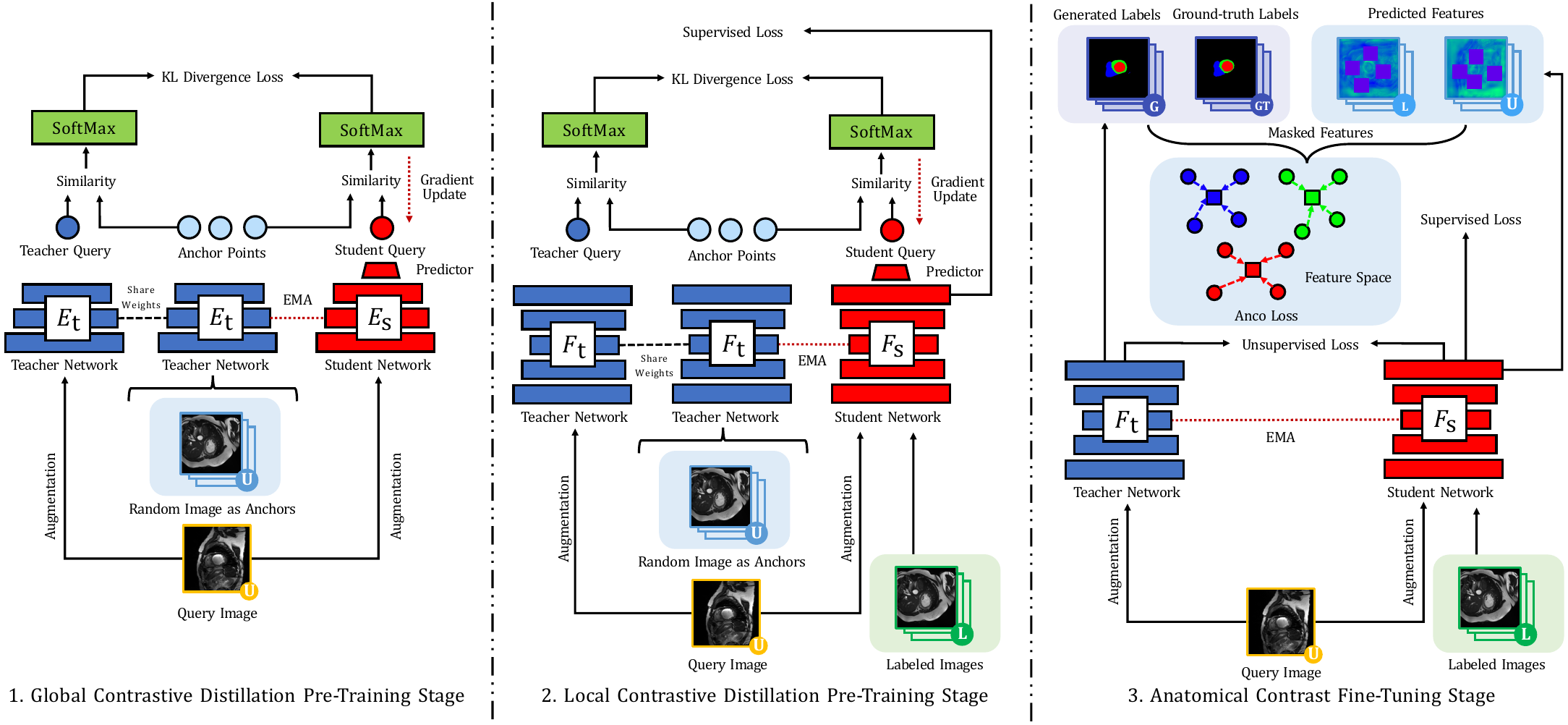}
\vspace{-20pt}
\caption{Overview of the ACTION framework including three stages: (1) global contrastive distillation pre-training used in existing works, (2) our proposed local contrastive distillation pre-training, and (3) our proposed anatomical contrast fine-tuning.} 
\label{fig:model}
\vspace{-10pt}
\end{figure}

\myparagraph{Framework Overview}
The workflow of our proposed ACTION is illustrated in Figure \ref{fig:model}. By default, ACTION is built on the BYOL pipeline \cite{grill2020bootstrap} which is originally designed for image classification tasks, and for a fair comparison, we also follow the setting in \cite{chaitanya2020contrastive} such as using 2D U-Net \cite{ronneberger2015u} as the backbone and non-linear projection heads $H$. The \tf{main differences} between our proposed ACTION and \cite{chaitanya2020contrastive,hu2021semi} are as follows: (1) the addition of a predictor $g(\cdot)$ to the student network to avoid collapsed solutions; (2) the utilization of a slow-moving average of the student network as the teacher network for more semantically compact representations; (3) the use of the output probability rather than logits effectively and semantically constrains the distance between the anatomical features from the imbalanced data (\ie, multi-class label imbalance cases); (4) we propose to contrast the query image features with other random image features at the global and local level, rather than only two augmented versions of the same image features; and (5) we design a novel unsupervised anatomical contrastive loss to provide additional supervision on hard pixels.

Let $(X, Y)$ be a training dataset including $N$ labeled image slices and $M$ unlabeled image slices, with training images $X\!=\!\{x_i\}_{i=1}^{N+M}$ and the $C$-class segmentation labels $Y\!=\!\{y_i\}_{i=1}^{N}$. Our backbone $F(\cdot)$ (2D U-Net) consists of an encoder network~$E(\cdot)$ and a decoder network $D(\cdot)$. The training procedure of ACTION includes three stages: ($i$) global contrastive distillation pre-training, ($ii$) local contrastive distillation pre-training, and ($iii$) anatomical contrast fine-tuning. In the first two stages, we use global contrastive distillation to train $E$ on unlabeled data to learn global-level features, and use local contrastive distillation to train $E$ and $D$ on labeled and unlabeled data to learn local-level features .

\myparagraph{Global Contrastive Distillation Pre-Training} 
We follow a similar setting in \cite{tejankar2021isd}. Given an input query image $q \in \{x_i\}_{i=N+1}^{N+M}$ with the spatial size $h\times w$, we first apply two different augmentations to obtain $q_{t}$ and $q_{s}$, and randomly sample a set of augmented images $\{x_j\}_{j=1}^{n}$ from a set of unlabeled image slices $\{x_i\}_{i=N+1}^{N+M}$. We believe that such relaxation enables the model to capture more rich semantic relationships and anatomical features from its neighboring images instead of only learning from the different version of the same query image. We then feed $\{x_j\}_{j=1}^{n}$ to the teacher encoder $E_t$, and followed by the nonlinear projection head $H_{t}^{g}$ to generate their projection embeddings $\{H_{t}^{g}(E_{t}(x_j))\}_{j=1}^{n}$ as anchor points, and also feed $q_{t}$ and $q_{s}$ to the teacher and student (\ie, $E$ and $H$), creating $z_t = H_{t}^{g}(E_{t}(q_{t}))$ and $z_s=H_{s}^{g}(E_{s}(q_{s}))$. Here we utilize the probabilities after SoftMax instead of the feature embedding:
\begin{equation}
p_t(j) = -\text{log} \frac{\text{exp}\big(\text{sim}\big(z_{t}, a_{j}\big)/\tau_t\big)}{\sum_{i=1}^n \text{exp}\big(\text{sim}\big(z_{t}, a_{i}\big)/\tau_t\big)},
\end{equation}
where $\tau_t$ is a temperature hyperparameter of the teacher, and $\text{sim}(\cdot,\cdot)$ is the cosine similarity. Then inspired by \cite{grill2020bootstrap}, in order to avoid collapsed solutions in an unsupervised scenario, we use a shallow multi-layer perceptron (MLP) predictor $H_{p}^{g}(\cdot)$ to obtain the prediction $z_{s}^{\ast}=H_{p}^{g}(z_{s})$. Of note, $\{a_{i}\}_{i=1}^{n}$, $z_{t}$, $z_{s}$, $z_{s}^{\ast}$ can be generated embedding from a set of randomly chosen augmented images, teacher's projection embeddings, student's projection embeddings, and student's prediction embeddings in either Stage-$i$ or $ii$. Therefore, we can calculate the similarity distance between the student's prediction and the anchor embeddings by converting them to probability distribution.
\begin{equation}
p_s(j) = -\text{log} \frac{\text{exp}\big(\text{sim}\big(z_{s}^{\ast}, a_{j}\big)/\tau_s\big)}{\sum_{i=1}^n \text{exp}\big(\text{sim}\big(z_{s}^{\ast}, a_{i}\big)/\tau_s\big)},
\end{equation}
where $\tau_s$ refers to a temperature hyperparameter of the student. The unsupervised contrastive loss is computed as follows:
\begin{equation}
\mathcal{L}_{\text{contrast}} = \text{KL}(p_t || p_s).
\label{equation:pcl}
\end{equation}
\myparagraph{Local Contrastive Distillation Pre-Training} 
After training the teacher's and student's encoder to learn global-level image features, we attach the decoders and tune the entire models to perform pixel-level contrastive learning in a semi-supervised manner. The \textbf{distinction in the training strategy} between ours and \cite{hu2021semi} lies in Stage-$ii$ and $iii$: \cite{hu2021semi} only use labeled data in training, while we use both labeled and unlabeled data in training. Considering the training procedure of Stage-$ii$ is similar to Stage-$iii$, we briefly describe it here as illustrated in Figure \ref{fig:model}. For the labeled data, we train our model by minimizing the supervised loss (the linear combination of cross-entropy loss and dice loss) in Stage-$ii$ and Stage-$iii$. As for the unlabeled input images $q$ and $\{x_j\}_{j=1}^{n}$, we first apply two different augmentations to $q$, creating two different versions $[q_{t}^{l},q_{s}^{l}]$, and then feed them to $F_{t}$ and $F_{s}$, and their output features $[f_{t},f_{s}]$ are fed into $H_{t}^{l}$ and $H_{t}^{l}$. The student's projection embedding is subsequently fed into $H_{p}^{l}$ to obtain the student's prediction embedding to enforce the similarity between the teacher and the student under the same loss as Equation \ref{equation:pcl}. We also include the randomly selected images to enforce such similarity because intuitively, it may be beneficial to ensure \tf{diversity} in the set of sampled examples. It is important to note that ACTION will re-use the well-trained weight of the models $F_t$ and $F_s$ as initialization for Stage-$iii$.

\myparagraph{Anatomical Contrast Fine-Tuning} Broadly speaking, in medical images, the same tissue types may share similar anatomical information in different patients, but different tissue types often show different class, appearance, and spatial distributions, which can be described as a complicated form of \tf{imbalance} and \tf{uncertainty} in real clinical data, as shown in Figure \ref{fig:distribution}. This motivates us to efficiently incorporate more useful features so the representations can be more balanced and better discriminated in such multi-class label imbalanced scenarios. Inspired by \cite{liu2021bootstrapping}, we propose AnCo, a new unsupervised contrastive loss designed at the anatomical level. Specifically, we additionally attach a representation decoder head $H_{r}$ to the student network, parallel to the segmentation head, to decode the multi-layer hidden features by first using multiple up-sampling layers for outputting dense features with the same spatial resolution as the query image and then mapping them into high $m$-dimensional query, positive key, and negative key embeddings: $r_q, r_k^{+}, r_k^{-}$. The AnCo loss is then defined as:
\begin{equation}
  \label{loss:anco}
   \mathcal{L}_\text{anco} = \sum_{c\in \mathcal{C}} \sum_{r_q \sim \mathcal{R}^c_q} -\log \frac{\exp(r_q \cdot r_k^{c, +} / \tau_{an})}{\exp(r_q \cdot r_k^{c, +}/ \tau_{an}) + \sum_{r_k^{-}\sim \mathcal{R}^c_k} \exp(r_q \cdot r_k^{-}/ \tau_{an})},
\end{equation}
where $\mathcal{C}$ is a set of all available classes in a mini-batch, and $\tau_{an}$ denotes a temperature hyperparameter for AnCo loss. $\mathcal{R}_q^c$ and $r_k^{c, +}$ are a set of query embeddings in class $c$ and the positive key embedding, which is the mean representation of class $c$, respectively. $\mathcal{R}_k^c$ is a set of negative key embeddings which are not in class $c$. Suppose $\mathcal{P}$ is a set including all pixel coordinates with the same resolution with $x_{i}$, these queries and keys are then defined as:
\begin{equation}
  \mathcal{R}_q^c\!=\!\!\bigcup_{[m, n]\in \mathcal{P}}\!\!\mathbbm{1}(y_{[m,n]}\!=\!c)\, r_{[m,n]},\, \mathcal{R}_k^c\!=\!\!\bigcup_{[m, n]\in \mathcal{P}}\!\!\mathbbm{1}(y_{[m,n]}\!\neq\!c)\, r_{[m,n]},\,
  r_k^{c, +}\!\!=\!\frac{1}{| \mathcal{R}_q^c |}\sum_{r_q \in \mathcal{R}_q^c} r_q.
\end{equation}
In addition, we note that contrastive learning usually benefits from a large collection of positive and negative pairs, but it is usually bounded by the size of GPU memory. Therefore, we introduce two novel active hard sampling methods. To address the \textit{uncertainty} on the most challenging pixels among all available classes (\ie, close anatomical or semantic relationship), we non-uniformly sample negative keys based on relative similarity distance between the query class and each negative key class. For each mini-batch, we build a graph $G$ to measure the pair-wise class relationship to dynamically update $G$.
\begin{equation}
  \label{eq:graph}
  G[p, q] = \left(r_k^{p, +} \cdot r_k^{q, +}\right),\quad  \forall p,q \in \mathcal{C}, \text{ and } p\neq q,
\end{equation}
where $G\in \mathbb{R}^{|\mathcal{C}| \times |\mathcal{C}|}$. Note that this process may be hard to allocate more samples. Thus, to learn a more accurate decision boundary, we first apply SoftMax function by normalizing the pair-wise relationships among all negative classes $n$ from each query class $c$, yielding a distribution: $\exp(G[c, v])/ \sum_{n\in \mathcal{C}, n\neq c} \exp(G[c, n])$. Then we adaptively sample negative keys from each class $v$ to help learn the corresponding query class $c$. To alleviate the \textit{imbalance} issue, we sample hard queries based on a defined threshold, to better discriminate the rare classes. The easy and hard queries are computed as follows:
\begin{equation}
  \label{eq: easyhard}
  \mathcal{R}_q^{c,\, easy} = \bigcup_{r_q \in \mathcal{R}^c_q} \mathbbm{1}(\hat{y}_q > \theta_s)r_q,\quad
  \mathcal{R}_q^{c,\, hard} = \bigcup_{r_q \in \mathcal{R}^c_q} \mathbbm{1}(\hat{y}_q \leq \theta_s)r_q,
\end{equation}
where $\hat{y}_q$ is the predicted confidence of label $c$ corresponding to $r_q$ after SoftMax function, and $\theta_s$ is the user-defined confidence threshold.
\section{Experiments}
\label{section:exp}
\myparagraph{Experimental Setup}
We experiment on two benchmark datasets: ACDC 2017 dataset \cite{bernard2018deep} and MICCAI 2017 Liver Tumor Segmentation Challenge (LiTS) \cite{bilic2019liver}. 

\noindent\textbf{The ACDC dataset} includes 200 cardiac cine MRI scans from 100 patients with annotations including three segmentation classes (\ie, left ventricle (LV), myocardium (Myo), and right ventricle (RV)). Following \cite{luo2020semi,wu2022mutual}, we use 140, 20, and 60 scans for training, validation, and testing, respectively.

\noindent\textbf{The LiTS dataset} includes 131 contrast-enhanced 3D abdominal CT volumes with annotations of two segmentation classes (\ie, liver and tumor). Following \cite{li2018h}, we use the first 100 volumes for training, and the rest 31 for testing. For pre-processing, we follow the setting in \cite{chaitanya2020contrastive} to normalize the intensity of each 3D scans, resample all 2D slices and the corresponding segmentation maps to a fixed spatial resolution (\ie, 256×256 pixels). To quantitatively assess the performance of our proposed method, we report two popular metrics: Dice coefficient (DSC) and Average Surface Distance (ASD) for 3D segmentation results.

\begin{table*}[t]
	\begin{center}
	\caption{Comparison of segmentation performance (DSC{[}\%{]}/ASD{[}voxel{]}) on ACDC under two unlabeled settings (3 or 7 labeled). The best results are indicated in \tf{bold}.}
	\vspace{-15pt}
	\label{table:acdc_main}
    \begin{adjustbox}{width=\linewidth}
	\begin{tabular}{ccccccccc}
		\toprule
		& & \multicolumn{3}{c}{3 Labeled} & & \multicolumn{3}{c}{7 Labeled} \\
        \cmidrule(r){3-5} \cmidrule(r){7-9}
		{Method}
		            & {Average}
		            & {RV}  
		            & {Myo}
		            & {LV}
		            & {Average}
		            & {RV}  
		            & {Myo}
		            & {LV} 
		            \\ \midrule
		UNet-F \cite{ronneberger2015u}
		            & {91.5}/{0.996} 
		            & {90.5}/{0.606}
                    & {88.8}/{0.941}
                    & {94.4}/{1.44}
		            & {91.5}/{0.996} 
		            & {90.5}/{0.606}
                    & {88.8}/{0.941}
                    & {94.4}/{1.44}
                    \\
		UNet-L        
                    & {51.7}/{13.1} 
		            & {36.9}/{30.1}
                    & {54.9}/{4.27}
                    & {63.4}/{5.11}
                    & {79.5}/{2.73}
		            & {65.9}/{0.892}
                    & {82.9}/{2.70}
                    & {89.6}/{4.60}
                    \\\midrule 
	    EM \cite{vu2019advent} 
                    & {59.8}/{5.64}
                    & {44.2}/{11.1}
                    & {63.2}/{3.23}
                    & {71.9}/{2.57}
                    & {75.7}/{2.73}
                    & {68.0}/{0.892}
                    & {76.5}/{2.70}
                    & {82.7}/{4.60}
                    \\ 
	    CCT \cite{ouali2020semi} 
                    & {59.1}/{10.1}
                    & {44.6}/{19.8}
                    & {63.2}/{6.04}
                    & {69.4}/{4.32}
                    & {75.9}/{3.60}
                    & {67.2}/{2.90}
                    & {77.5}/{3.32}
                    & {82.9}/{0.734}
                    \\ 
	    DAN \cite{zhang2017deep} 
                    & {56.4}/{15.1}
                    & {47.1}/{21.7}
                    & {58.1}/{11.6}
                    & {63.9}/{11.9}
                    & {76.5}/{3.01}
                    & {75.7}/{2.61}
                    & {73.3}/{3.11}
                    & {80.5}/{3.31}
                    \\
	    URPC \cite{luo2021efficient} 
                    & {58.9}/{8.14}
                    & {50.1}/{12.6}
                    & {60.8}/{4.10}
                    & {65.8}/{7.71}
                    & {73.2}/{2.68}
                    & {67.0}/{0.742}
                    & {72.2}/{0.505}
                    & {80.4}/{6.79}
                    \\ 
	    DCT \cite{qiao2018deep} 
                    & {58.5}/{10.8}
                    & {41.2}/{21.4}
                    & {63.9}/{5.01}
                    & {70.5}/{6.05}
                    & {78.1}/{2.64}
                    & {70.7}/{1.75}
                    & {77.7}/{2.90}
                    & {85.8}/{3.26}
                    \\ 
	    ICT \cite{verma2019interpolation} 
                    & {59.0}/{6.59}
                    & {48.8}/{11.4}
                    & {61.4}/{4.59}
                    & {66.6}/{3.82}
                    & {80.6}/{1.64}
                    & {75.1}/{0.898}
                    & {80.2}/{1.53}
                    & {86.6}/{2.48}
                    \\ 
	    MT \cite{tarvainen2017mean} 
                    & {58.3}/{11.2}
                    & {39.0}/{21.5}
                    & {58.7}/{7.47}
                    & {77.3}/{4.72}
                    & {80.1}/{2.33}
                    & {75.2}/{1.22}
                    & {79.2}/{2.32}
                    & {86.0}/{3.45}
                    \\ 
	    UAMT \cite{yu2019uncertainty} 
                    & {61.0}/{7.03}
                    & {47.8}/{15.9}
                    & {65.0}/{2.38}
                    & {70.1}/{2.83}
                    & {77.6}/{3.15}
                    & {70.5}/{0.81}
                    & {78.4}/{4.36}
                    & {83.9}/{4.29}
                    \\ 
	    CPS \cite{chen2021semi} 
                    & {61.0}/{2.92}
                    & {43.8}/{2.95}
                    & {64.5}/{2.84}
                    & {74.8}/{2.95}
                    & {78.8}/{3.41}
                    & {74.0}/{1.95}
                    & {78.1}/{3.11}
                    & {84.5}/{5.18}
                    \\ 
	    GCL \cite{chaitanya2020contrastive} 
                    & {70.6}/{2.24}
                    & {56.5}/{1.99}
                    & {70.7}/{1.67}
                    & {84.8}/{3.05}
                    & {87.0}/{0.751}
                    & {86.9}/\tf{0.584}
                    & {81.8}/{0.821}
                    & {92.5}/{0.849}
                    \\ 
	    SCS \cite{hu2021semi} 
                    & {73.6}/{5.37}
                    & {63.5}/{6.23}
                    & {76.6}/{2.42}
                    & {80.7}/{7.45}
                    & {84.2}/{2.01}
                    & {81.4}/{0.850}
                    & {83.0}/{2.03}
                    & {88.2}/{3.12}
                    \\ 
        \gr \cb ACTION (ours)
                    & \tf{87.5}/\tf{1.12}
                    & \tf{85.4}/\tf{0.915}
                    & \tf{85.8}/\tf{0.784}
                    & \tf{91.2}/\tf{1.66}
                    & \tf{89.7}/\tf{0.736}
                    & \tf{89.8}/{0.589}
                    & \tf{86.7}/\tf{0.813}
                    & \tf{92.7}/\tf{0.804}
                    \\ 
		              \bottomrule
	\end{tabular}
    \end{adjustbox}
    \end{center}
    \vspace{-20pt}
\end{table*}
\begin{figure}[ht]
\centering
\includegraphics[width=0.95\linewidth]{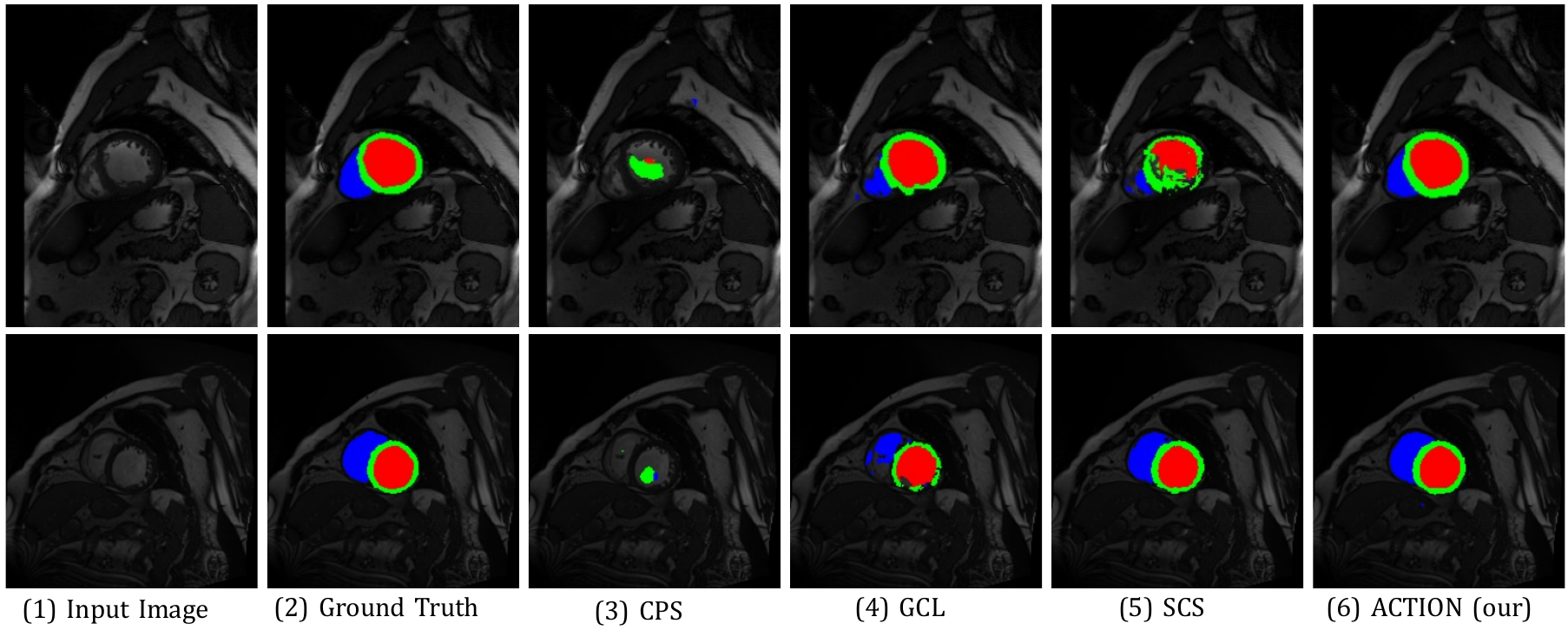}
\vspace{-10pt}
\caption{Visualization of segmentation results on ACDC with 3 labeled data. As is shown, ACTION consistently produces sharper object boundaries and more accurate predictions across all methods. Different structure categories are shown in different colors.} 
\label{fig:vis_acdc}
\vspace{-15pt}
\end{figure}

\begin{table*}[ht]
	\begin{center}
	\caption{Comparison of segmentation performance (DSC{[}\%{]}/ASD{[}voxel{]}) on LiTS under two unlabeled settings (5\% or 10\% labeled ratio). The best results are in \tf{bold}.}
	\vspace{-10pt}
	\label{table:lits_main}
    \begin{adjustbox}{width=0.9\linewidth}
	\begin{tabular}{ccccccc}
		\toprule
		& & \multicolumn{2}{c}{5\% Labeled} & & \multicolumn{2}{c}{10\% Labeled} \\
        \cmidrule(r){3-4} \cmidrule(r){6-7}
		{Method}
		            & {Average}
		            & {Liver}  
		            & {Tumor}
		            & {Average}
		            & {Liver}  
		            & {Tumor}
		            \\ \midrule
		UNet-F \cite{ronneberger2015u}
		            & {68.2}/{16.9} 
		            & {90.6}/{8.14}
                    & {45.8}/{25.6}
		            & {68.2}/{16.9} 
		            & {90.6}/{8.14}
                    & {45.8}/{25.6}
                    \\
		UNet-L        
                    & {60.4}/{30.4} 
		            & {87.5}/{9.84}
                    & {33.3}/{50.9}
                    & {61.6}/{28.3} 
		            & {85.4}/{18.6}
                    & {37.9}/{37.9}
                    \\\midrule 
	    EM \cite{vu2019advent} 
                    & {61.2}/{33.3}
                    & {87.7}/{9.47}
                    & {34.7}/{57.1}
                    & {62.9}/{38.5}
                    & {87.4}/{21.3}
                    & {38.3}/{55.7}
                    \\ 
                    
	    CCT \cite{ouali2020semi} 
                    & {60.6}/{48.7}
                    & {85.5}/{27.9}
                    & {35.6}/{69.4}
                    & {63.8}/{31.2}
                    & {90.3}/{7.25}
                    & {37.2}/{55.1}
                    \\ 
	    DAN \cite{zhang2017deep} 
                    & {62.3}/{25.8}
                    & {88.6}/{9.64}
                    & {36.1}/{42.1}
                    & {63.2}/{30.7}
                    & {87.3}/{15.4}
                    & {39.1}/{46.1}
                    \\
	    URPC \cite{luo2021efficient} 
                    & {62.4}/{37.8}
                    & {86.7}/{21.6}
                    & {38.0}/{54.0}
                    & {63.0}/{43.1}
                    & {88.1}/{24.3}
                    & {38.9}/{61.9}
                    \\ 
	    DCT \cite{qiao2018deep} 
                    & {60.8}/{34.4}
                    & {89.2}/{12.6}
                    & {32.5}/{56.2}
                    & {61.9}/{31.7}
                    & {86.2}/{19.3}
                    & {37.5}/{44.1}
                    \\ 
	    ICT \cite{verma2019interpolation} 
                    & {60.1}/{39.1}
                    & {86.8}/{12.6}
                    & {33.3}/{65.6}
                    & {62.5}/{32.4}
                    & {88.1}/{16.7}
                    & {36.9}/{48.2}
                    \\ 
	    MT \cite{tarvainen2017mean} 
                    & {61.9}/{40.0}
                    & {86.7}/{21.6}
                    & {37.2}/{58.4}
                    & {63.3}/{26.2}
                    & {89.7}/{11.6}
                    & {36.9}/{40.8}
                    \\ 
	    UAMT \cite{yu2019uncertainty} 
                    & {61.0}/{47.0}
                    & {86.9}/{22.1}
                    & {35.2}/{71.8}
                    & {62.3}/{26.0}
                    & {87.4}/{7.55}
                    & {37.3}/{44.4}
                    \\ 
	    CPS \cite{chen2021semi} 
                    & {62.1}/{36.0}
                    & {87.3}/{17.9}
                    & {36.8}/{54.0}
                    & {64.0}/{23.6}
                    & {90.2}/{10.6}
                    & {37.8}/{36.7}
                    \\ 
	    GCL \cite{chaitanya2020contrastive} 
                    & {63.3}/{20.1}
                    & {90.7}/{9.46}
                    & {35.9}/{30.8}
                    & {65.0}/{37.2}
                    & {91.3}/{10.0}
                    & {38.7}/{64.3}
                    \\ 
	    SCS \cite{hu2021semi} 
                    & {61.5}/{28.8}
                    & {92.6}/{7.21}
                    & {30.4}/{50.3}
                    & {64.6}/{33.9}
                    & {91.6}/{5.72}
                    & {37.6}/{62.0}
                    \\ 
        \gr \cb ACTION (ours)
                    & \tf{66.8}/\tf{17.7}
                    & \tf{93.0}/\tf{6.04}
                    & \tf{40.5}/\tf{29.4}
                    & \tf{67.7}/\tf{20.4}
                    & \tf{92.8}/\tf{5.08}
                    & \tf{42.6}/\tf{35.8}
                    \\ 
		              \bottomrule
	\end{tabular}
    \end{adjustbox}
    \end{center}
    \vspace{-10pt}
\end{table*}

\begin{figure}[h]
\centering
\includegraphics[width=0.95\linewidth]{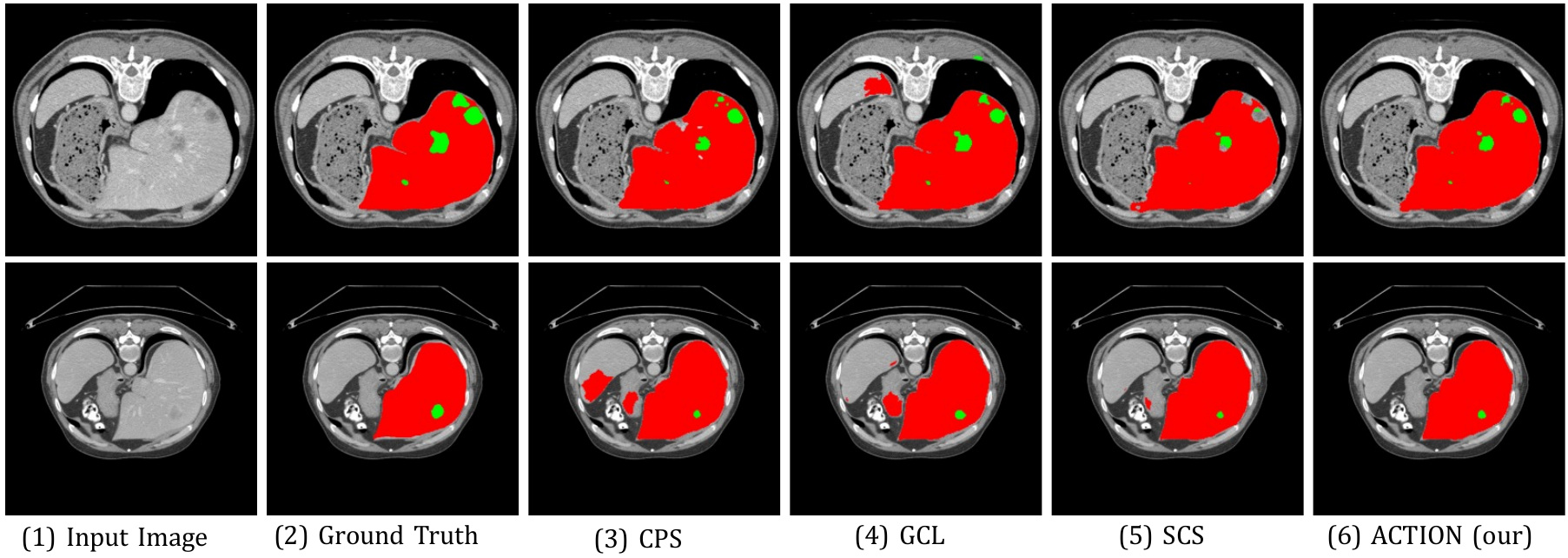}
\vspace{-10pt}
\caption{Visualization of segmentation results on LiTS with 5\% labeled ratio. As is shown, ACTION achieves consistently sharp and accurate object boundaries compared to other SSL methods. Different structure categories are shown in different colors.} 
\label{fig:vis_lits}
\end{figure}

\myparagraph{Implementation Details}
All our models are implemented in PyTorch \cite{paszke2019pytorch}. We train all methods with SGD optimizer (learning rate=$0.01$, momentum=$0.9$, weight decay=$0.0001$, batch size=$6$). All models are trained with two NVIDIA GeForce RTX 3090 GPUs. Stage-$i$ and $ii$ are trained with 100 epochs, and Stage-$iii$ is with 200 epochs. We use the temperature of teacher and student as $\tau_t\!=\!0.01$ and $\tau_s\!=\!0.1$. The teacher is updated using the following rule $ \theta_t \leftarrow m \theta_t + (1-m) \theta_s$, where $\theta$ refers to the model's parameters and the the momentum hyperparameter $m$ is $0.99$. The memory bank size is 36. We follow the standard augmentation strategies in \cite{grill2020bootstrap}. In Stage-$i$, we train $E_{s}$, $E_{t}$, $H_{t}^{g}$, $H_{s}^{g}$, and $H_{p}^{g}$ on the unlabeled data with global-level $\mathcal{L}_{\text{contrast}}$ in Equation 3. We follow \cite{hu2021semi} to use a MLP as heads, and the setting of the predictors is similar to \cite{grill2020bootstrap}, which has a feature dimension of $512$. In Stage-$ii$, we train $F_{s}$, $F_{t}$, $H_{t}^{l}$, $H_{s}^{l}$, and $H_{p}^{l}$ on the labeled and unlabeled data. We train with the supervised loss \cite{yu2019uncertainty} on labeled data, and local-level $\mathcal{L}_{\text{contrast}}$ in Equation 3 on unlabeled data. Given the logits output $\hat{y} \in \mathbb{R}^{C\times h\times w}$, we use the $1\times 1$ convolutional layer to project all pixels into the latent space with the feature dimension of $512$, and the output feature dimension of $G$ is also $512$. As for Stage-$iii$, we train $F_{s}$, $F_{t}$, $H_{t}$, $H_{s}$, and $H_{r}$ on the labeled and unlabeled data. We use the supervised segmentation loss on labeled data, unsupervised cross-entropy loss (on pseudo-labels generated by a confidence threshold $\theta_s$), and $\mathcal{L}_{\text{anco}}$ in Equation 4 on unlabeled data. We then adaptively sample 256 query samples and 512 key samples for each mini-batch, and temperature for the student and confidence thresholds are set to $\tau_{s}=0.5$ and $\theta_s=0.97$, respectively. Of note, the projection heads, the predictor, and the representation decoder head are only utilized during the training, and will be removed during the inference.

\myparagraph{Main Results}
We compare our proposed method to previous state-of-the-art SSL methods using 2D Unet \cite{ronneberger2015u} as backbone, including UNet trained with full/limited supervisions (UNet-F/UNet-L), EM \cite{vu2019advent}, CCT \cite{ouali2020semi}, DAN \cite{zhang2017deep}, URPC \cite{luo2021efficient}, DCT \cite{qiao2018deep}, ICT \cite{verma2019interpolation}, MT \cite{tarvainen2017mean}, UAMT \cite{yu2019uncertainty}, CPS \cite{chen2021semi}, SCS \cite{hu2021semi}, and GCL \cite{chaitanya2020contrastive}. 
Table \ref{table:acdc_main} shows the evaluation results on ACDC dataset under two unlabeled settings (3 or 7 labeled cases). ACTION can substantially improve results on two unlabeled settings, greatly outperforming the previous state-of-the-art SSL methods. Specifically, our ACTION, trained on 3 labeled cases, dramatically improves the previous best averaged Dice score from 73.6\% to 87.5\% by a large margin, and even matches previous SSL methods using 7 labeled cases. When using 7 labeled cases, ACTION further pushes the state-of-the-art results to 89.7\% in Dice. We observe that the gains are more pronounced on the two categories(\ie, RV and Myo), and our ACTION achieves 89.8\% and 86.7\% in terms of Dice, performing competitive or even better than the supervised baseline (89.2\% and 86.7\%). As shown in Figure \ref{fig:vis_acdc}, we can see the clear advantage of ACTION, where the boundaries of different regions are clearly sharper and more accurate such as RV and Myo regions. Table \ref{table:lits_main} also shows the evaluation results on LiTS dataset under two unlabeled settings (5\% or 10\% labeled cases). On both two labeled settings, ACTION significantly outperforms all the state-of-the-art methods by a significant margin. As shown in Figure \ref{fig:vis_lits}, ACTION achieves consistently sharp and accurate object boundaries compared to other SSL methods.

\begin{table}[t]
\centering
\caption{Ablation on {\bf (a)} model component: w/o Random Sampled Images (RSI); w/o Local Contrastive Distillation (Stage-$ii$); w/o Anatomical Contrast Fine-tuning (Stage-$iii$); {\bf (b)} loss formulation: w/o $\mathcal{L}_\mathrm{anco}$; w/o $\mathcal{L}_\mathrm{unsup}$;, compared to the Vanilla and our proposed ACTION. Note that $\mathcal{L}_\mathrm{unsup}$ denotes cross-entropy loss (on pseudo-labels generated by a confidence threshold $\theta_s$) together with $\mathcal{L}_\mathrm{anco}$ used in Stage-$iii$.} 
\vspace{-5pt}
\label{tab:component_ablation}
\resizebox{0.5\textwidth}{!}{
\begin{tabular}{c|l|c c} 
\toprule
& \,\multirow{2}{*}{Method} & \multicolumn{2}{c}{\textbf{Metrics}} \\ 
    & & Dice{[}\%{]}
    & ASD{[}voxel{]} 
    \\ \midrule
    & \,Vanilla  
    & 60.6
    & 6.64
    \\
	& \,ACTION (ours)
    & \textbf{87.5}
    & \textbf{1.12}   
    \\ 
    \midrule
    \multirow{4}{*}{\bf (a)} 
    & \quad w/o RSI
    & 82.7
    & 6.66
    \\
    & \quad w/o Stage-$ii$
    & 86.4
    & 1.69
    \\
    & \quad w/o RSI + Stage-$ii$ \quad
    & 82.6
    & 1.77
    \\
    & 
    \quad w/o Stage-$iii$
    & 76.7
    & 2.91
    \\ \midrule
    \multirow{2}{*}{\bf (b)}
    & \quad w/o $\mathcal{L}_\mathrm{anco}$
    & 86.5
    & 1.30
    \\
    &   
    \quad w/o $\mathcal{L}_\mathrm{unsup}$ 
    & 83.7
    & 2.51
    \\ 
    \bottomrule
\end{tabular}
}
\end{table}


\myparagraph{Ablation on Different Components}
We investigate the impact of different components in ACTION. All reported results in this section are based on the ACDC dataset under the 3 labeled setting.
Table \ref{tab:component_ablation} shows the ablation result of our model. Upon our choice of architecture, we first consider a na\"ive baseline (BYOL) without any random sampled images (RSI), stage-$ii$, and stage-$iii$, denoted by ({1}) Vanilla. Then, we consider a wide range of different settings for improved representation learning: (2) incorporating other random sampled images; (3) no stage-$ii$; (4) no other random sampled images and stage-$ii$; (5) no stage-$iii$; since stage-$iii$ includes two losses, (6) no $\mathcal{L}_\mathrm{anco}$, (7) no $\mathcal{L}_\mathrm{unsup}$, and (8) our proposed ACTION. As shown in Table \ref{tab:component_ablation}, it is notable that ACTION performs generally better than other evaluated baselines. We find that only applying any single component of ACTION often comes at the cost of performance degradation. The intuitions behind are as follows: (1) incorporating other random sampled images will enforce the diversity of the sampled data, preventing redundant anatomically and semantically similar samples; (2) using stage-$ii$ leads to worse performance without considering local context; (3) using stage-$iii$ enables a robust segmentation model to learn better representations with few human annotations. Using the above components confers a significant advantage at representation learning, and further illustrates the benefit of each component.

\begin{table}[t]
\centering
\caption{Ablation on augmentation strategies.} 
\vspace{-5pt}
\label{tab:aug_ablation}
\resizebox{0.5\textwidth}{!}{
\begin{tabular}{ll|c c|c c} 
\toprule
&\multirow{2}{*}{Method}  
&\multicolumn{2}{c|}{Student Teacher}
&\multicolumn{2}{c}{\textbf{Metrics}} \\ 
& & \multicolumn{2}{c|}{Aug.  Aug.} & Dice{[}\%{]} & ASD{[}voxel{]}
    \\ \midrule
    \gr &  ACTION  
    & Weak
    & Weak
    & 84.6
    & 1.78
    \\
	& ACTION
    & Strong
    & Weak
    & 87.5
    & 1.12
    \\ 
    \gr &  ACTION
    & Weak
    & Strong
    & 85.4
    & 2.12
    \\  
    & ACTION
    & Strong
    & Strong
    & 86.5
    & 1.89
    \\ 
    \bottomrule
\end{tabular}
}
\end{table}


\myparagraph{Ablation on Different Augmentations}
We investigate the impact of using weak or strong augmentations for ACTION on the ACDC dataset under 3 labeled setting. We summarize the effects of different data augmentation strategies in Table \ref{tab:aug_ablation}. We apply \textit{weak} augmentation to the teacher's input, including rotation, cropping, flipping, and \textit{strong} augmentation to the student's input, including rotation, cropping, flipping, random contrast, and brightness changes \cite{perez2018data}. Empirically, we find that when using weak and strong augmentation strategies on the teacher and student network, the network performance is optimal.
\section{Conclusion and Limitations}
In this work, we have presented ACTION, a novel anatomical-aware contrastive distillation framework with active sampling, designed specifically for medical image segmentation. Our method is motivated by two observations that all negative samples are not equally negative, and the underlying class distribution of medical images is highly unlabeled and imbalanced. Through extensive experiments across two benchmark datasets and unlabeled settings, we show that ACTION can significantly improve segmentation performance with minimal additional memory requirements, outperforming the previous state-of-the-art by a large margin. For future work, we plan to explore a more advanced contrastive learning approach for better performance when the medical data is unlabeled and imbalanced.

%
%
%
\bibliographystyle{splncs04}
\bibliography{ref}

\end{document}